# Repeated Robot-Assisted Unilateral Stiffness Perturbations Result in Significant Aftereffects Relevant to Post-Stroke Gait Rehabilitation

Vaughn Chambers and Panagiotis Artemiadis*, *IEEE Senior Member*

*Abstract*— Due to hemiparesis, stroke survivors frequently develop a dysfunctional gait that is often characterized by an overall decrease in walking speed and a unilateral decrease in step length. With millions currently affected by this dysfunctional gait, robust and effective rehabilitation protocols are needed. Although robotic devices have been used in numerous rehabilitation protocols for gait, the lack of significant aftereffects that translate to effective therapy makes their application still questionable. This paper proposes a novel type of robot-assisted intervention that results in significant aftereffects that last much longer than any other previous study. With the utilization of a novel robotic device, the Variable Stiffness Treadmill (VST), the stiffness of the walking surface underneath one leg is decreased for a number of steps. This unilateral stiffness perturbation results in a significant aftereffect that is both useful for stroke rehabilitation and often lasts for over 200 gait cycles after the intervention has concluded. More specifically, the aftereffect created is an increase in both left and right step lengths, with the unperturbed step length increasing significantly more than the perturbed. These effects may be helpful in correcting two of the most common issues in post-stroke gait: overall decrease in walking speed and a unilateral shortened step length. The results of this work show that a robot-assisted therapy protocol involving repeated unilateral stiffness perturbations can lead to a more permanent and effective solution to post-stroke gait.

## I. INTRODUCTION

Stroke is a severe problem currently affecting an estimated 7 million Americans [1]. Those who survive a stroke are often left with a motor impairment affecting their gait [2]. This is mainly due to hemiparesis, which is the paralysis or partial paralysis of one side of the body. The most common issues hemiparetic walkers face are an overall decrease in walking speed and a unilateral shortened step length [3].

Post-stroke rehabilitation is still an important healthcare challenge as current practice is not extremely effective [4]. While robots have generally shown to be a promising and useful rehabilitation tool primarily on upper limbs [5], [6], this has not yet translated to post-stroke gait rehabilitation [7]–[15]. Body-weight-supported treadmills have been widely used and standardized for rehabilitation of gait impairments [16], [17]. The state of the art devices for gait rehabilitation impose gait kinematics on the impaired legs using either *hard* or *soft* means, ranging from kinematically controlled exoskeletons [10], [18]–[20] to impedance-controlled orthotic devices [12], [21]–[24]. According to recent studies, there is moderate evidence of improvement in walking and motor recovery using robotic devices including systems for body-weight supported treadmill training (BWSTT) when compared to conventional therapy [14]–[16], [24]–[28].

In order for robot-assisted therapy to promote motor recovery and rehabilitation, it should elicit specific aftereffects related to improved gait after the intervention is over. Along these lines, there have been studies using split-belt treadmills where a meaningful aftereffect was shown [29], [30]. This was done in one particular study through the principle of error augmentation by setting the treadmill belts to different speeds for 10-minute long intervals. After having the subjects walk on this setting, they then resume walking in an unperturbed environment. It was observed that subjects improved their step length and double limb support during this period. This effect lasted for 25 gait cycles [29]. While this study showed that unilateral perturbations could evoke adaptations in human gait, a longer lasting aftereffect is desired for effective gait rehabilitation.

Our group has developed a unique robotic device, the Variable Stiffness Treadmill (VST) [31], [32], which has been used in many studies for understanding and improving human gait [33]–[39]. In a nutshell, the VST is a split-belt treadmill that has the capability of dynamically altering the vertical stiffness (or compliance) of the left belt, as humans walk on it. This change is unilateral, i.e. the two belts of the treadmill are independent. Using the capabilities of the VST we have previously shown that unilateral stiffness perturbations evoke an immediate response in the contralateral leg. The latency between the perturbation and the contralateral response has been shown to be greater than 150ms, therefore suggesting that the brain is involved in this process [40]. This has been verified in a later study with brain recordings [41]. These results of brain involvement are promising as the concept of stroke rehabilitation through neuroplasticity most importantly requires repeated active involvement of the damaged area of the brain [42], [43].

Although we have shown promising results during and immediately following perturbed gait cycles, never before have we attempted to show that unilateral stiffness perturbations result in a lasting aftereffect. In this paper, we attempt for the first time to systematically elicit and characterize aftereffects with specific functional outcomes on gait, which are caused by a repeated set of unilateral stiffness perturbations. More specifically, we repeatedly perturb the left leg of subjects by lowering the walking surface stiffness using the VST. After having the subjects walk on the unilaterally lowered stiffness

*This material is based upon work supported by the National Science Foundation under Grants No. #2020009, #2015786, #2025797, and #2018905.

Vaughn Chambers and Panagiotis Artemiadis are with the Mechanical Engineering Department, at the University of Delaware, Newark, DE 19716, USA. vaughn@udel.edu, partem@udel.edu

*Corresponding author: partem@udel.edu

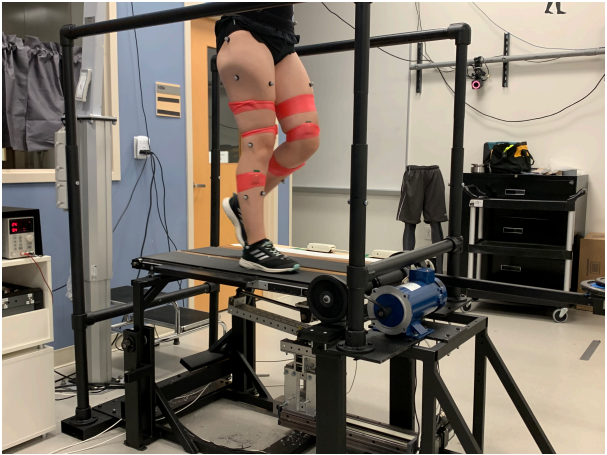

Fig. 1. Subject walking on VST with motion capture markers, EMG sensors, and harness partly shown on subject's waist/torso.

environment for 400 gait cycles, we bring the walking conditions back to normal (rigid treadmill on both sides). We show for the first time that unilateral stiffness perturbations result in an aftereffect once the perturbations are removed that is useful for post-stroke gait rehabilitation. The aftereffects at a functional level are an increased step length for both legs, with the right (unperturbed) leg increasing significantly more (3.07% increase) than the left (2.28% increase). These effects last more than 200 gait cycles in a majority of our subjects which is significantly more than what has been shown in previous studies. These findings are very promising for post-stroke gait rehabilitation, as they directly relate to some of the most common issues seen in stroke patients, such as decreased walking speed and unilateral reduced step length.

The remainder of this paper is organized as follows: Section II introduces the methods and experimental protocol used with intact subjects on the VST, as well as the data collection and processing. Section III presents the results of our study, focusing on the specific aftereffects on gait and their main causes from a kinematic and muscular activation point of view. Section IV concludes the paper by pointing out the contribution of the work and possible applications and directions for future work.

## II. METHODS

The main robotic device used in this work is the Variable Stiffness Treadmill (VST) (shown in Fig. 1). The VST is a split-belt treadmill that has the capability of altering the vertical stiffness of the left belt independently from the right belt. The achievable stiffness range is from about 60 N/m to 1 MN/m, the latter of which is considered to be rigid. This device allows for subjects to walk in an environment where one leg is experiencing a ground stiffness level that is less than rigid, while the other remains rigid. This allows for the creation of a variety of environments that perturb the subject unilaterally. To relate our treadmill to the real world, walking on the VST can be comparable to walking with one foot on sand and the other on pavement. For more information on the VST, see previous works where its design and capabilities have been described in detail [31], [32].

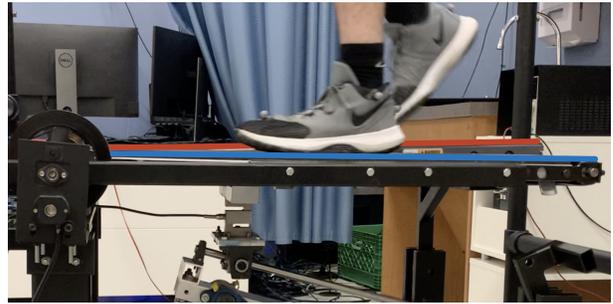

Fig. 2. Deflection of the left VST belt at a stiffness of 45 kN/m. The left belt (blue) and the right belt (red) are highlighted for clarity. The subject's weight is 85 kg for reference.

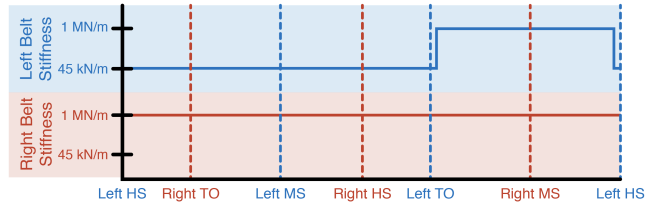

Fig. 3. Timing of unilateral stiffness perturbation within one gait cycle. Note that: HS, MS, and TO stand for heel strike, mid-stance, and toe-off respectively. The stiffness is adjusted to rigid during the swing phase, when the leg is not touching the treadmill, to avoid oscillations of the belt after the leg pushes off.

### A. Experimental Protocol

Eight healthy subjects partook in this study (age: $23.8 \pm 1.3$ years, height: $169.5 \pm 9.7$ cm, weight: $68.5 \pm 12.2$ kg). All subjects were free from neurological or musculoskeletal disorders that affect their ability to walk and balance. The subjects were asked to walk on the VST using the protocol outlined below. Informed consent was given, while these experimental protocols are approved by the University of Delaware Institutional Review Board (IRB ID#: 1544521-2).

For this experiment, only two stiffness values are used for the left belt: 1 MN/m (rigid) and 45 kN/m. For a frame of reference, 45 kN/m feels similarly to walking on sand or a soft gym/yoga mat. In Fig. 2, the amount of deflection created by a 85 kg human subject walking on the VST at the instance of left single support, at a stiffness level of 45 kN/m, is shown. This value of 45 kN/m was chosen based on preliminary testing to avoid excessive vertical displacement on the left side and subject fatigue. The right belt of the treadmill remains rigid for the entirety of this experiment. For this study, the stiffness remains constant throughout the left stance phase. In other words, there are only two kinds of gait cycles: those that are rigid (unperturbed) and those where the stiffness of the left belt of the treadmill is at 45 kN/m for the entire left stance phase (perturbed). During the left swing phase, the left treadmill belt returned to rigid to reduce oscillations after the leg push-off. Note that this of course does not affect the subject as they are not in contact with the left belt of the treadmill during left swing phase. The timing of a perturbed gait cycle can be seen in Fig. 3. Note that the perturbation ends just after stance phase ends, and starts just before stance phase starts. This was done simply to ensure that the left leg is always experiencing a stiffness of 45 kN/m on a perturbed gait cycle.

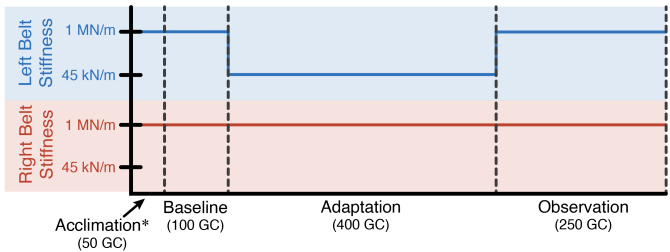

Fig. 4. Layout of experiment. The total experiment is 800 gait cycles, where a gait cycle (GC) consists of a full step with each leg (from left heel strike to the next left heel strike). The graphs show the stiffness level of each belt (left and right) during the stance phase of each leg. *The acclimation phase was not used in any data analysis as its sole purpose was to get the subjects accustomed to walking on our treadmill.

The total duration of the experiment was 800 gait cycles and contained four sections: acclimation, baseline, adaptation, and observation (see Fig. 4). The acclimation phase of the experiment lasted for 50 gait cycles, and the treadmill remained rigid for this entire period. No data from this section was used, as its sole purpose was to introduce the subject to walking on our treadmill. The baseline phase lasted for 100 gait cycles, and the treadmill was also rigid for this phase. The adaptation period lasted for 400 gait cycles, and the left side of the treadmill was set to 45 kN/m during every left stance phase. During this section, as well as the entire experiment, the right side of the treadmill remained rigid. Last, the observation phase lasted for 250 gait cycles, and the treadmill remained rigid for the entirety of this section.

Each subject was equipped with 22 passive motion capture markers attached to their lower body used for tracking the motion of their legs. The markers were tracked with 8 VICON cameras recording at a frequency of 100 Hz. Eight wireless surface electromyographic (EMG) electrodes (Trigno, Delsys Inc.) were used for recording muscular activity of major muscles of the lower limbs: tibialis anterior (TA), gastrocnemius (GA), vastus medialis (VA), and biceps femoris (BF) of each leg. The EMG signals were sampled at 2 kHz and synchronized with the motion capture data. A body weight support harness was used during this experiment, but did not offset any weight. This harness went around the subject's torso and was only used for safety precautions. (see Fig. 1).

The subjects were asked to choose their self-selected walking speed before data collection began. To achieve this, the subjects walked on the VST and told the experimenter to either speed up or slow down the treadmill. The treadmill speed was increased or decreased in increments of 5 cm/s. This was done until the subject found a walking speed that they felt was a neither leisurely nor hurried. For all 8 subjects, the chosen speed was in the range of 80 to 90 cm/s.

Before the start of the experiment, the subjects were instructed to keep their arms above their hips to ensure that all motion capture markers were visible. This was achieved by either keeping their elbows fairly flexed as they swung their arms or by resting the backs of their hands on the handrails. Note that the latter instruction was given to keep them from using the handrails as a major balancing tool. The subjects were also told that if they felt unsafe at anytime they could hold on to the handrails or ask to stop the experiment, but neither of these ever occurred.

*B. Data Processing*

When processing the synchronized raw kinematic and muscular activity data, our newly developed algorithm, the F-VESPA algorithm, was used to detect heel strikes [44], [45]. From this, the data was segregated into gait cycles using the left heel strike as the starting and ending point for each gait cycle (see Fig. 3). Outlier gait cycles were identified at this point based on their length in time. On average a total of 32 ± 10 outlier gait cycle were found per subject and excluded from the total data (800 gait cycles). Also, using the detected heel strikes and toe-off events, left and right stance and swing phases were able to be identified for the entire experiment.

To investigate the presence of an aftereffect once the stiffness was set back to rigid, the gait cycles during the observation phase were incrementally compared to the data of the baseline phase. This was done by separating the observation phase into groups of 10 gait cycles. To begin, the first group of 10 gait cycles was compared to the entire baseline phase using a statistical significance test (see below). If the result of this test was deemed to be significant, the first group of 10 and the second group of 10 were combined to make a larger group that was then compared to the baseline phase. If this was also deemed significant, the group would increase to the first 30 gait cycles, and so on. This was done until the there were either no more groups of 10 gait cycles left to compare, or a test returned a non-significant result. We believe this method for significance testing aligns with the goal of this study of finding out how long an aftereffect lasts, while allowing for slight variability across steps inherent to human gait [46].

In all significance tests, the Wilcoxon rank-sum test (also known as the Mann-Whitney U test) was used with an $\alpha$ value of 0.05 [47]. This test was chosen due to its non-parametric nature, making it more robust with the small sample sizes that occur when checking for an aftereffect in the first few observation gait cycles groups, as described in the paragraph above.

EMG data were processed using the following method. The raw data were first filtered with a 4th-order Butterworth band-pass filter with low and high frequencies at 30 Hz and 300 Hz, respectively. The data were then full-wave rectified. The envelope of the signal was then found using the moving mean value with a window of 400 data points (200 ms). Last, the data were filtered with a 4th-order Butterworth low-pass filter with a cut-off frequency of 5 Hz. The final processed EMG data were normalized based on their maximum values found throughout the experiment for each muscle.

Last, all data were filtered one more time before displaying the figures included in the paper using the Robust Loess quadratic fit (RLOESS) method with a span of 250 samples. This last filtering step was done after significance testing and does not affect the results of the study. The RLOESS filtering was done only to allow trends to be seen more clearly.

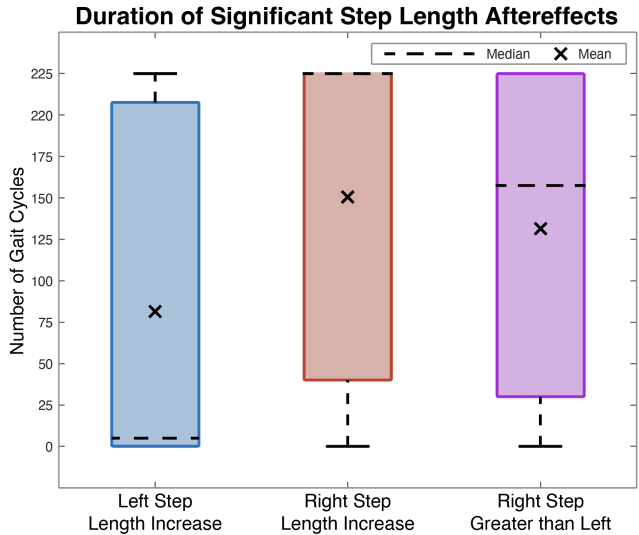

Fig. 5. Box and whisker plot summarizing the data from all eight subjects. The left plot shows for how many gait cycles after the perturbations ended the left step length remained increased. The center plot shows for how many gait cycles after the perturbations ended the right step length remained increased. The right plot shows for how many gait cycles after the perturbations ended the right step length was significantly greater than the left. Note that the black horizontal lines are the median value and the "x" marks the mean value. Significance is considered valid when the statistical significance test (Wilcoxon rank-sum test) results in a probability less than $\alpha = 0.05$. Note that this graph ends at 225 gait cycles to account for outlier gait cycles. Therefore a significant aftereffect lasting 225 gait cycles is synonymous with significance lasting until the end of the experiment.

## III. Results

Figure notation: All figures in this section use the following notation: blue colors correspond to the left side of the body, red colors correspond to the right side of the body and purple colors correspond to the comparison between left and right sides.

The results of this study show that repeated unilateral stiffness perturbations result in a larger step length for both legs, with the contralateral (unperturbed, right) step length increasing significantly more than the ipsilateral (perturbed, left). On average, the left step length increased by 1.09 cm (2.28%) and this increase lasted for 81 gait cycles, and the right step length increased by 1.47 cm (3.07%) and lasted for 150 gait cycles, across all subjects. Note that the step length in this study was defined as the anterior-posterior distance between left and right ankles upon heel strike. Left and right heel strike are distinguished based on which foot is in front (left heel strike → left step length, right heel strike → right step length). The overall results for all eight subjects are summarized in Fig. 5. This figure depicts for how long each aftereffect remained significant (using the protocol outlined in the Methods section). As stated in the Methods section, significance is considered valid when the Wilcoxon rank-sum test results in a p-value less than $\alpha = 0.05$. As can be seen, the mean values of all three parameters are above 75 gait cycles. Most notably, the median value for the right step length increase is 225 gait cycles, meaning that over half of the subjects experienced a significant increase in right step length for the entire observation phase of the experiment.

### A. Averaged Data

While each subject responded differently to the perturbations, we decided to average the data of all eight subjects in an attempt to display the big picture trends and reduce the variability of individual subjects. This was done by taking the average of each data point seen on Fig. 6, 7, and 8. In other words, the parameter of interest is averaged across all eight subjects, for every gait cycle. Note that certain subjects had more outlier gait cycles than others, basically shortening their data set. To account for this when averaging across subjects, the shortest baseline, adaptation, and observation phases were used. These outliers are why all graphs stop before the 800 gait cycle length of the experiment.

After averaging the data, we found the following results. The right step length was significantly increased for the entire observation phase (Fig. 6). Figure 6 shows the right step length for each gait cycle throughout the entire experiment, while the three phases are noted. A significance test comparing the observation gait cycles with the baseline gait cycles, as described in Methods, is implemented, and when the statistical test showed a significant difference ($p<0.05$), a significance line is added to the figure. It can be seen that the adaptation phase causes the right step length to increase significantly. Once the observation phase starts, and the treadmill returns to rigid, the affect of increased step length remains, and it is statistically significant until the end of the experiment. Moreover, even though the experiment ended at 700 gait cycles, the effect did not appear to be diminishing. Next, the averaged difference in step length can be observed in Fig. 7. In this graph, the right step length is subtracted from the left step length for each gait cycle of the experiment. As denoted by the significance line in the bottom right of the figure, this trend remained significant for the entire observation phase as well. It can be seen that while step lengths in the baseline phase were equal (as expected), they are asymmetric in the observation phase, with the right step length being greater than the left. While unlike the averaged right step length, we can observe this effect diminishing and returning to baseline. An aftereffect is still very much present though.

It should be noted at this point that we can be assured these aftereffects are due to the unilateral stiffness perturbations and not just an effect of prolonged treadmill walking. Control trials have been run on the VST where a subject simply walks in an unperturbed environment for 800 gait cycles. Those trials have shown no significant changes in step length or any other parameter discussed in this paper.

To illustrate the mechanisms at play to achieve such an important functional outcome in gait, we look first at the kinematics. The increase in step length was mainly due to an increase in hip flexion during late swing and heel strike (seen in 7 out of 8 subjects). This can be seen graphically in Fig. 8. Figure 8 shows a very clear increase in the hip flexion angle at heel strike for the left leg, and this increase again stays statistically significant for the entirety of the observation phase. Note that similar trends were seen with the right leg as well. Additionally, 3 out of 8 subjects also experienced greater knee extension during late swing and heel strike (not shown in a figure). This is also believed to contribute to a larger step length, but since this was not seen in the majority of subjects we will not discuss it any further.

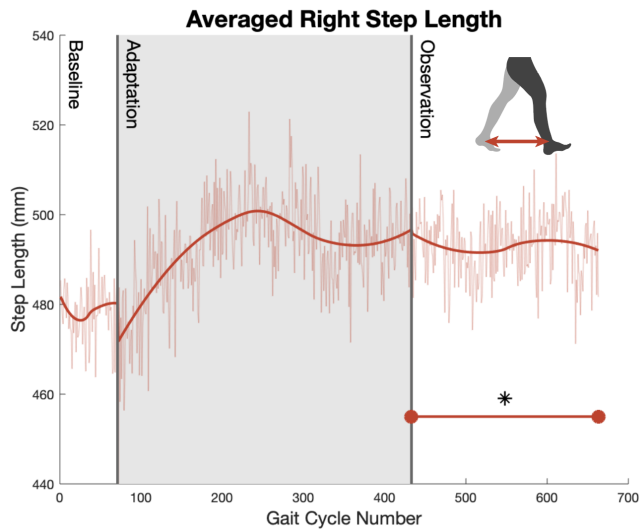

Fig. 6. Averaged right step length for all eight subjects. The solid line represents smoothed data and the thin line represents averaged data before the RLOESS filtering (see Methods for more information). The right step length remains significantly increased for the entire observation phase with an average value of +14.7mm above the average baseline value.

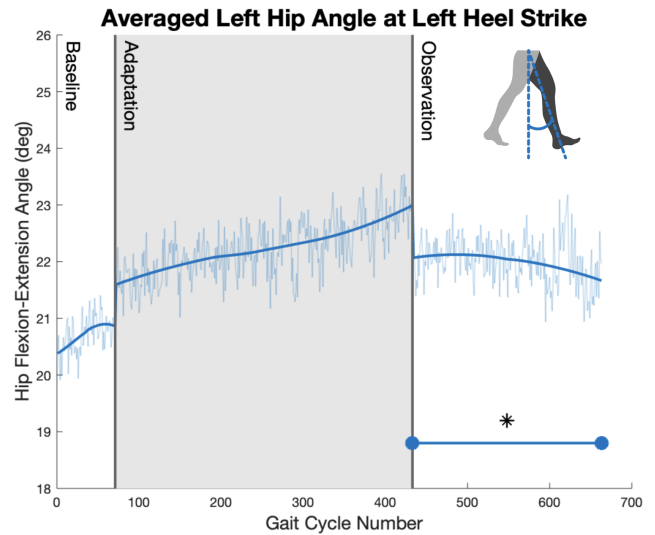

Fig. 8. The left hip flexion-extension angle at each left heel strike, averaged across all eight subjects. The solid line represents smoothed data and the thin line represents averaged data before the RLOESS filtering (see Methods for more information). The significant increase in hip flexion is believed to be the main cause of the observed increased step length.

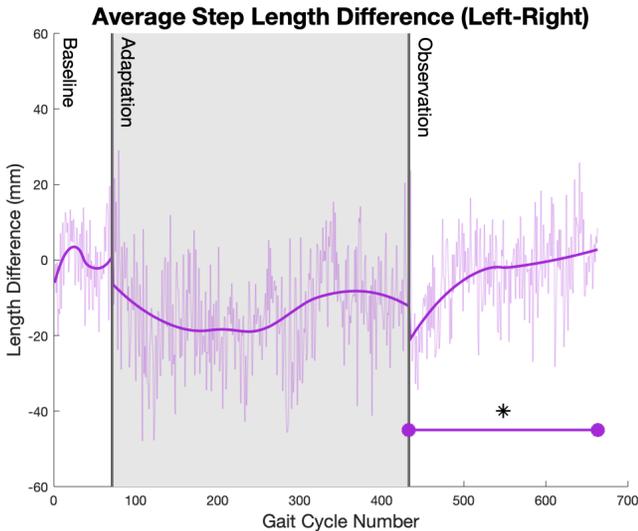

Fig. 7. Averaged difference in step length (left-right) for all eight subjects. The solid line represents smoothed data and the thin line represents averaged data before the RLOESS filtering (see Methods for more information). The right step length remains significantly more increased than the left for the entire observation phase.

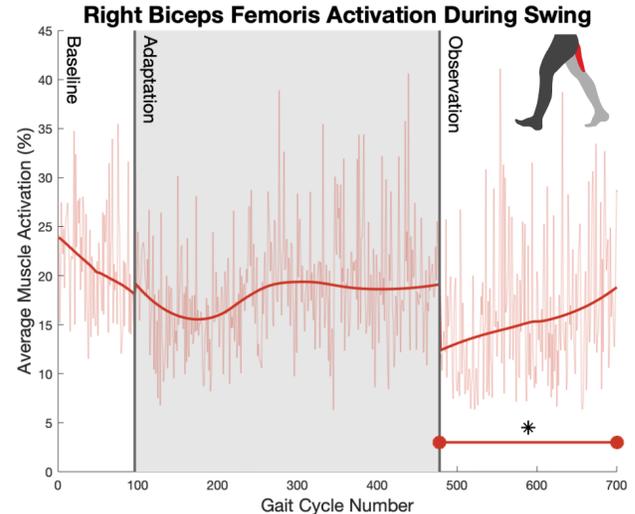

Fig. 9. The average right biceps femoris activation for each right swing phase. Data from a representative subject are shown. The solid line represents smoothed data and the thin line represents averaged data before the RLOESS filtering (see Methods for more information). The activation remains statistically significantly less than that of the baseline for the whole duration of the observation phase of the experiment as noted by the significance line in the bottom right of the figure.

### B. Representative Subject

Note that since EMG data are notoriously noisy, we will only be looking at muscle activity data from a representative subject in this section.

To explain the increase in hip flexion and subsequent increase in step length through muscle activity, one needs to first consider the muscles acting about the hip. As muscles around the hip act in an agonist-antagonist manner, we expect that increased flexion would be caused by increased activity on the hip flexor muscles, and decreased activity on the hip extensor muscles, during swing phase. Our results show that the biceps femoris (hip extensor) displays a clear aftereffect mainly described by decreased muscle activity during the swing phase of the right leg, allowing for greater hip flexion (see Fig. 9). The average activity of the muscle in the swing phase is shown. The biceps femoris activation remains statistically significantly less than that of the baseline for the whole duration of the observation phase of the experiment. Note that the left biceps femoris showed similar results.

Increased hip flexion can also be explained through greater activation of the gastrocnemius and vastus medialis muscles.

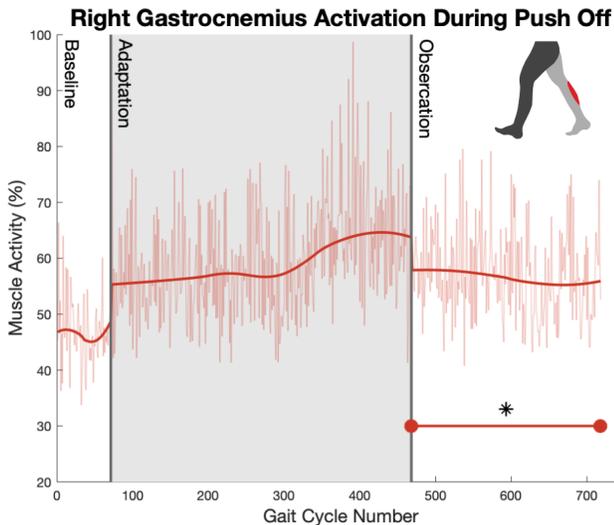

Fig. 10. The right gastrocnemius activation for each right push off. Note that this graph does not display average values as the other EMGs figures do, but rather show the maximum muscle activity from late stance to early swing. Data from a representative subject are shown. The solid line represents smoothed data and the thin line represents averaged data before the RLOESS filtering (see Methods for more information). The activation remains statistically significantly greater than that of the baseline for the whole duration of the observation phase of the experiment as noted by the significance line in the bottom right.

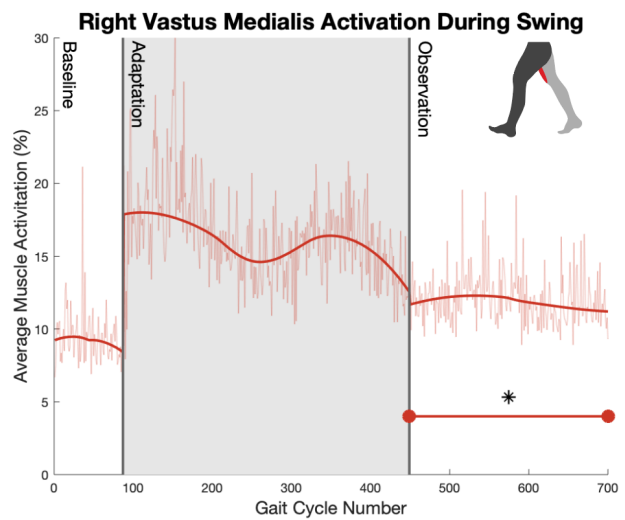

Fig. 11. The average right vastus medialis activation for each right swing phase. Data from a representative subject are shown. The solid line represents smoothed data and the thin line represents averaged data before the RLOESS filtering (see Methods for more information). The activation remains statistically significantly greater than that of the baseline for the whole duration of the observation phase of the experiment as noted by the significance line in the bottom right.

First, an increased gastrocnemius activation during push-off can assist in propelling the leg further forward, leading to greater hip flexion at heel strike. This can be seen in Fig. 10, with an aftereffect lasting for the entire observation phase. This result was seen in four out of eight subjects. In a similar manner, increased vastus medialis activation during swing can also propel the leg forward. If the lower portion of the leg (shank) is moving forward faster, this will add extra momentum to the leg, therefore leading to more hip flexion in late swing and at heel strike. This result can be seen in Fig. 11. This aftereffect lasts the entire observation phase and was seen in five out of eight subjects. Even though both of these results are shown for the right leg, the left leg showed similar results.

To summarize, our results provide strong evidence that our unique type of perturbation results in prolonged aftereffects with a specific functional outcome in gait, which is evoked by specific changes in muscle activation.

## IV. CONCLUSIONS

As stated before, very common issues with post-stroke gait are an overall decrease in walking speed and unilateral shortened step length. The results of this study may help in correcting both of these issues by causing an aftereffect of increased step length on both legs, with the unperturbed leg increasing significantly more than the perturbed. This aftereffect not only appears to be functionally relevant to post-stroke gait rehabilitation, but in many cases it lasts for the remainder of the experiment (200+ gait cycles). An aftereffect of this length has yet to be seen in existing post-stroke gait rehabilitation methods. This shows that the unique unilateral stiffness perturbation created on the VST can possibly elicit adaptations in the neural and musculoskeletal mechanisms responsible for the control of human walking, with specific implications for the rehabilitation of gait.

We believe the effects seen after repeated unilateral stiffness perturbations are so significant due to the way they disrupt interlimb coordination. This coordination is perturbed via the following neurological pathways, which have all been shown to be commonly damaged after stroke: proprioception [48], [49], force feedback [50], and balance [51]. Having the capability of disrupting these three pathways is what makes the VST so unique. Other gait rehabilitation devices (exoskeletons, standard split-belt treadmills) are not able to perturb the subject's gait in such a way.

To conclude, this paper suggests for the first time that after repeated unilateral stiffness perturbations, the subject carries over an aftereffect to subsequent unperturbed walking that often lasts for 200 gait cycles. Functionally this aftereffect is increased step length of both the perturbed and unperturbed leg, where the unperturbed leg increases significantly more than the perturbed. This is accomplished mainly through increased hip flexion and increased gastrocnemius and vastus medialis muscle activity. These findings are very promising as they relate to some of the most common issues seen in stroke patients: decreased walking speed and unilateral reduced step length. The presence of an aftereffect lasting over 200 gait cycles shows that the nervous system is capable of adapting and storing new ways of accomplishing a task such as unperturbed, steady-state walking, given the appropriate perturbations. Based on our results, we believe that the main contribution of this paper is the introduction and analysis of a therapy protocol involving repeated unilateral stiffness perturbations, which can lead to more permanent adaptations and therefore more effective robot-assisted gait rehabilitation than what has been achieved thus far.

## References

[1] E. J. Benjamin *et al.*, "Heart Disease and Stroke Statistics-2019 Update: A Report From the American Heart Association," *Circulation*, vol. 139, no. 10, pp. e56–e528, mar 2019.

[2] P. W. Duncan, R. Zorowitz, B. Bates, J. Y. Choi, J. J. Glasberg, G. D. Graham, R. C. Katz, K. Lamberty, and D. Reker, "Management of Adult Stroke Rehabilitation Care," *Stroke*, vol. 36, no. 9, sep 2005.

[3] S. Li, G. E. Francisco, and P. Zhou, "Post-stroke hemiplegic gait: New perspective and insights," *Frontiers in Physiology*, vol. 9, no. AUG, p. 1021, aug 2018.

[4] A. R. Carter, L. T. Connor, and A. W. Dromerick, "Rehabilitation after stroke: current state of the science," *Current neurology and neuroscience reports*, vol. 10, no. 3, pp. 158–166, 2010.

[5] C. G. Burgar, P. S. Lum, P. C. Shor, and H. F. M. Van der Loos, "Development of robots for rehabilitation therapy: the Palo Alto VA/Stanford experience," *Journal of rehabilitation research and development*, vol. 37, no. 6, pp. 663–674, 2000.

[6] J. Stein, R. Hughes, S. Fasoli, H. I. Krebs, and N. Hogan, "Clinical applications of robots in rehabilitation," *Critical Reviews™ in Physical and Rehabilitation Medicine*, vol. 17, no. 3, 2005.

[7] D. R. Louie and J. J. Eng, "Powered robotic exoskeletons in post-stroke rehabilitation of gait: a scoping review," *Journal of neuroengineering and rehabilitation*, vol. 13, no. 1, pp. 1–10, 2016.

[8] S. Jezernik, G. Colombo, T. Keller, H. Frueh, and M. Morari, "Robotic Orthosis Lokomat: A Rehabilitation and Research Tool," *Neuromodulation*, vol. 6, no. 2, pp. 108–115, apr 2003.

[9] S. Hesse, D. Uhlenbrock, C. Werner, and A. Bardeleben, "A mechanized gait trainer for restoring gait in nonambulatory subjects," *Archives of Physical Medicine and Rehabilitation*, vol. 81, no. 9, pp. 1158–1161, sep 2000.

[10] J. F. Veneman, R. Kruidhof, E. E. Hekman, R. Ekkelenkamp, E. H. Van Asseldonk, and H. Van Der Kooij, "Design and evaluation of the LOPES exoskeleton robot for interactive gait rehabilitation," *IEEE Transactions on Neural Systems and Rehabilitation Engineering*, vol. 15, no. 3, pp. 379–386, sep 2007.

[11] A. Morbi, M. Ahmadi, and A. Nativ, "GaitEnable: An omnidirectional robotic system for gait rehabilitation," in *2012 IEEE International Conference on Mechatronics and Automation, ICMA 2012*, 2012, pp. 936–941.

[12] S. K. Banala, S. K. Agrawal, and J. P. Scholz, "Active Leg Exoskeleton (ALEX) for gait rehabilitation of motor-impaired patients," in *2007 IEEE 10th International Conference on Rehabilitation Robotics, ICORR'07*, 2007, pp. 401–407.

[13] M. Peshkin, D. A. Brown, J. J. Santos-Munné, A. Makhlin, E. Lewis, J. E. Colgate, J. Patton, and D. Schwandt, "KineAssist: A robotic overground gait and balance training device," in *Proceedings of the 2005 IEEE 9th International Conference on Rehabilitation Robotics*, vol. 2005, 2005, pp. 241–246.

[14] T. G. Hornby, D. D. Campbell, J. H. Kahn, T. Demott, J. L. Moore, and H. R. Roth, "Enhanced gait-related improvements after therapist- versus robotic-assisted locomotor training in subjects with chronic stroke: a randomized controlled study." *Stroke*, vol. 39, no. 6, pp. 1786–92, jun 2008.

[15] J. Hidler, D. Nichols, M. Pelliccio, K. Brady, D. D. Campbell, J. H. Kahn, and T. G. Hornby, "Multicenter randomized clinical trial evaluating the effectiveness of the Lokomat in subacute stroke," *Neurorehabilitation and Neural Repair*, vol. 23, no. 1, pp. 5–13, jan 2009.

[16] B. H and V. M, "Optimal outcomes obtained with body-weight support combined with treadmill training in stroke subjects," *Archives of physical medicine and rehabilitation*, vol. 84, pp. 1458–1465, 2003.

[17] S. Hesse, C. Bertelt, M. T. Jahnke, A. Schaffrin, P. Baake, M. Malezic, and K. H. Mauritz, "Treadmill training with partial body weight support compared with physiotherapy in nonambulatory hemiparetic patients." *Stroke*, vol. 26(6), pp. 976–981, 1995.

[18] D. Ferris, J. Czerniecki, and B. Hannaford, "An ankle-foot orthosis powered by artificial pneumatic muscles," in *Journal of Applied Biomechanics*, 2005, pp. 189–197.

[19] G. Colombo, M. Joerg, and R. Schreier, "Treadmill training of paraplegic patients using a robotic orthosis," *J. Rehabil. Res. Dev.*, vol. 37, pp. 693–700, 2000.

[20] M. Pohl, C. Werner, M. Holzgraefe, G. Kroczek, J. Mehrholz, I. Wingendorf, G. Hoolig, R. Koch, and S. Hesse, "Repetitive locomotor training and physiotherapy improve walking and basic activities of daily living after stroke: a single-blind, randomized multicentre trial (Deutsche Gang trainer Studie, DEGAS)," *Clinical Rehabilitation*, vol. 21, pp. 17–27, 2007.

[21] A. Roy, H. I. Krebs, D. Williams, C. T. Bever, L. W. Forrester, R. M. Macko, and N. Hogan, "Robot-aided neurorehabilitation: A robot for ankle rehabilitation," *IEEE Transaction on Robotics*, vol. 25:3, pp. 569–582, 2009.

[22] J. A. Blaya and H. Herr, "Adaptive control of a variable-impedance ankle-foot orthosis to assist drop-foot gait," *Neural Systems and Rehabilitation Engineering, IEEE Transactions on*, vol. 12, no. 1, pp. 24–31, 2004.

[23] H. I. Krebs, N. Hogan, M. L. Aisen, and B. T. Volpe, "Robot-aided neurorehabilitation," *IEEE Transactions on Rehabilitation Engineering*, vol. 6:1, pp. 75–87, 1998.

[24] R. Riener, L. Lunenburger, S. Jezernik, M. Anderschitz, G. Colombo, and V. Dietz, "Patient-cooperative strategies for robot-aided treadmill training: first experimental results," *Neural Systems and Rehabilitation Engineering, IEEE Transactions on*, vol. 13, no. 3, pp. 380–394, 2005.

[25] A. Mayr, M. Kofler, E. Quirbach, H. Matzak, K. Fröhlich, and L. Saltuari, "Prospective, blinded, randomized crossover study of gait rehabilitation in stroke patients using the lokomat gait orthosis," *Neurorehabilitation and Neural Repair*, vol. 21, pp. 307–314, 2007.

[26] A. R. Luft *et al.*, "Treadmill exercise activates subcortical neural networks and improves walking after stroke: a randomized controlled trial," *Stroke*, vol. 39, no. 12, pp. 3341–3350, 2008.

[27] B. Bates, J. Y. Choi, P. W. Duncan, J. J. Glasberg, G. D. Graham, R. C. Katz, K. Lamberty, D. Reker, and R. Zorowitz, "Veterans Affairs/Department of Defense Clinical Practice Guideline for the Management of Adult Stroke Rehabilitation Care Executive Summary," *Stroke*, vol. 36, no. 9, pp. 2049–2056, 2005.

[28] C. J. Winstein *et al.*, "Guidelines for adult stroke rehabilitation and recovery a guideline for healthcare professionals from the american heart association/american stroke association," *Stroke*, vol. 47, no. 6, pp. e98–e169, 2016.

[29] D. S. Reisman, R. Wityk, K. Silver, and A. J. Bastian, "Split-Belt Treadmill Adaptation Transfers to Overground Walking in Persons Poststroke," *Neurorehabilitation and Neural Repair*, vol. 23, pp. 735–744, 2009.

[30] K. V. Huynh, C. H. Sarmento, R. T. Roemmich, E. L. Stegemöller, and C. J. Hass, "Comparing aftereffects after split-belt treadmill walking and unilateral stepping," *Medicine and Science in Sports and Exercise*, vol. 46, no. 7, pp. 1392–1399, 2014.

[31] A. Barkan, J. Skidmore, and P. Artemiadis, "Variable Stiffness Treadmill (VST): A novel tool for the investigation of gait," in *Proceedings - IEEE International Conference on Robotics and Automation*. Institute of Electrical and Electronics Engineers Inc., sep 2014, pp. 2838–2843.

[32] J. Skidmore, A. Barkan, and P. Artemiadis, "Variable Stiffness Treadmill (VST): System Development, Characterization, and Preliminary Experiments," *IEEE/ASME Transactions on Mechatronics*, vol. 20, no. 4, pp. 1717–1724, aug 2015.

[33] J. Skidmore and P. Artemiadis, "Unilateral floor stiffness perturbations systematically evoke contralateral leg muscle responses: a new approach to robot-assisted gait therapy," *IEEE Transactions on Neural Systems and Rehabilitation Engineering*, vol. 24, no. 4, pp. 467–474, 2016.

[34] R. Frost, J. Skidmore, M. Santello, and P. Artemiadis, "Sensorimotor control of gait: a novel approach for the study of the interplay of visual and proprioceptive feedback," *Frontiers in human neuroscience*, vol. 9, 2015.

[35] J. Skidmore and P. Artemiadis, "Sudden changes in walking surface compliance evoke contralateral emg in a hemiparetic walker: A case study of inter-leg coordination after neurological injury," in *Engineering in Medicine and Biology Society (EMBC), 2016 IEEE 38th Annual International Conference of the*. IEEE, 2016, pp. 4682–4685.

[36] J. Skidmore and P. Artemiadis, "On the effect of walking surface stiffness on inter-limb coordination in human walking: toward bilaterally informed robotic gait rehabilitation," *Journal of neuroengineering and rehabilitation*, vol. 13, no. 1, p. 32, 2016.

[37] J. Skidmore and P. Artemiadis, "Leg muscle activation evoked by floor stiffness perturbations: A novel approach to robot-assisted gait rehabilitation," in *Robotics and Automation (ICRA), 2015 IEEE International Conference on*. IEEE, 2015, pp. 6463–6468.

[38] J. Skidmore and P. Artemiadis, "Unilateral changes in walking surface compliance evoke dorsiflexion in paretic leg of impaired walkers," *Journal of Rehabilitation and Assistive Technologies Engineering*, vol. 4, p. 2055668317738469, 2017.

[39] J. Skidmore and P. Artemiadis, "A comprehensive analysis of sensorimotor mechanisms of inter-leg coordination in gait using the variable stiffness treadmill: Physiological insights for improved robot-assisted gait therapy," in *2019 IEEE 16th International Conference on Rehabilitation Robotics (ICORR)*. IEEE, 2019, pp. 28–33.

[40] J. Skidmore and P. Artemiadis, "Unilateral Floor Stiffness Perturbations Systematically Evoke Contralateral Leg Muscle Responses: A New Approach to Robot-Assisted Gait Therapy," *IEEE Transactions on Neural Systems and Rehabilitation Engineering*, vol. 24, no. 4, pp. 467–474, apr 2016.

[41] J. Skidmore and P. Artemiadis, "Unilateral walking surface stiffness perturbations evoke brain responses: Toward bilaterally informed robot-assisted gait rehabilitation," in *Proceedings - IEEE International Conference on Robotics and Automation*, vol. 2016-June. IEEE, may 2016, pp. 3698–3703.

[42] Y. R. S. Su, A. Veeravagu, and G. Grant, "Neuroplasticity after traumatic brain injury," in *Translational Research in Traumatic Brain Injury*. CRC Press, apr 2016, ch. 8, pp. 163–178.

[43] J. J. Daly and R. L. Ruff, "Construction of Efficacious Gait and Upper Limb Functional Interventions Based on Brain Plasticity Evidence and Model-Based Measures For Stroke Patients," *Discussion Paper TheScientificWorldJOURNAL*, vol. 7, pp. 2031–2045, 2007.

[44] C. Karakasis and P. Artemiadis, "Real-time kinematic-based detection of foot-strike during walking," *Journal of Biomechanics*, vol. 129, p. 110849, 2021.

[45] C. Karakasis and P. Artemiadis, "F-vespa: A kinematic-based algorithm for real-time heel-strike detection during walking," in *2021 IEEE/RSJ International Conference on Intelligent Robots and Systems (IROS)*. IEEE, 2021, pp. 5098–5103.

[46] N. Stergiou, *Nonlinear analysis for human movement variability*. CRC press, 2018.

[47] W. Haynes, *Wilcoxon Rank Sum Test*. New York, NY: Springer New York, 2013, pp. 2354–2355.

[48] D. Rand, "Proprioception deficits in chronic stroke—Upper extremity function and daily living," *PLOS ONE*, vol. 13, no. 3, p. e0195043, mar 2018.

[49] J. C. Tuthill and E. Azim, "Proprioception," *Current Biology*, vol. 28, no. 5, pp. R194–R203, 2018.

[50] L. M. Carey, T. A. Matyas, and L. E. Oke, "Sensory loss in stroke patients: Effective training of tactile and proprioceptive discrimination," *Archives of Physical Medicine and Rehabilitation*, vol. 74, no. 6, pp. 602–611, 1993.

[51] S. F. Tyson and L. H. DeSouza, "Reliability and validity of functional balance tests post stroke," *Clinical Rehabilitation*, vol. 18, no. 8, pp. 916–923, 2004.